\documentclass[conference]{IEEEtran}
\IEEEoverridecommandlockouts
\usepackage[T1]{fontenc}
\usepackage{cite}
\usepackage{amsmath,amssymb,amsfonts}
\usepackage{algorithmic}
\usepackage{graphicx}
\usepackage{textcomp}
\usepackage{xcolor}
\usepackage{algorithm}
\usepackage{algorithmic}
\usepackage{amsmath}

\DeclareUnicodeCharacter{0307}{\.{}}
\DeclareUnicodeCharacter{2212}{-}

\def\BibTeX{{\rm B\kern-.05em{\sc i\kern-.025em b}\kern-.08em
    T\kern-.1667em\lower.7ex\hbox{E}\kern-.125emX}}
\begin{document}

\title{Revolutionizing Glioma Segmentation \& Grading Using 3D MRI - Guided Hybrid Deep Learning Models\\
}

\author{
\IEEEauthorblockN{
Pandiyaraju V\IEEEauthorrefmark{1},
Sreya Mynampati\IEEEauthorrefmark{2},
Abishek Karthik\IEEEauthorrefmark{3},
Poovarasan L\IEEEauthorrefmark{4},
D. Saraswathi\IEEEauthorrefmark{5}
}

\IEEEauthorblockA{
\IEEEauthorrefmark{1,2,3,5}\textit{Department of Computer Science and Engineering, School of Computer Science and Engineering,}\\
\textit{Vellore Institute of Technology, Chennai, India}
}

\IEEEauthorblockA{
\IEEEauthorrefmark{4}\textit{School of Advanced Science, Vellore Institute of Technology, Chennai, India}
}

\IEEEauthorblockA{
\IEEEauthorrefmark{5}\textit{Centre for Human Movement Analytics, Vellore Institute of Technology, Chennai, India}
}

\IEEEauthorblockA{
\IEEEauthorrefmark{1}pandiyaraju.v@vit.ac.in \quad
\IEEEauthorrefmark{2}sreyamynampati@gmail.com \quad
\IEEEauthorrefmark{3}abishek.sudhirkarthik@gmail.com \\
\IEEEauthorrefmark{4}poovarasan.l2024@vitstudent.ac.in \quad
\IEEEauthorrefmark{5}saraswathi.d@vit.ac.in
}
}

\maketitle

\begin{abstract}
Gliomas are brain tumor types that have a high mortality rate which means early and accurate diagnosis is important for therapeutic intervention for the tumors. Traditionally, however, assessing Magnetic Resonance Imaging (MRI) has been a labor intensive process and prone to human error and ultimately provides limited value of information for tumors. To address this difficulty, the proposed research will develop a hybrid deep learning model which integrates U-Net based segmentation and a hybrid DenseNet-VGG classification network with multi-head attention and spatial-channel attention capabilities. The segmentation model will precisely demarcate the tumors in a 3D volume of MRI data guided by spatial and contextual information. The classification network which combines a branch of both DenseNet and VGG, will incorporate the demarcated tumor on which features with attention mechanisms would be focused on clinically relevant features.

The primary goals aim to establish a suitable system for glioma detection and grading that minimizes manual diagnostic errors and inspection time as well as optimizes the model to find the rapid clinical outcome. High-dimensional 3D MRI data could successfully be utilized in the model through preprocessing steps which are normalization, resampling, and data augmentation. Through a variety of measures the framework is evaluated: measures of performance in segmentation are Dice coefficient and Mean Intersection over Union (IoU) and measures of performance in classification are accuracy precision, recall, and F1-score.

The hybrid framework that has been proposed has demonstrated through physical testing that it has the capability of obtaining a Dice coefficient of 98\% in tumor segmentation, and 99\% on classification accuracy, outperforming traditional CNN models and attention-free methods. Utilizing multi-head attention mechanisms enhances notions of priority in aspects of the tumor that are clinically significant, and enhances interpretability and accuracy. The results suggest a great potential of the framework in facilitating the timely and reliable diagnosis and grading of glioma by clinicians is promising, allowing for better planning of patient treatment.

\end{abstract}


\section{Introduction}

Gliomas are the most frequently occurring primary malignant brain tumors. They constitute about 81\% of all malignant intracranial tumors and are a significant public health burden with high clinical and socioeconomic value. These aggressive tumors arise from glial cells within the central nervous system, primarily the brain, and exhibit a high degree of heterogeneity with respect to histological, growth, and clinical characteristics. The World Health Organization (WHO) classifies gliomas into grades I to IV based on histological characteristics and malignant behavior. High-Grade Gliomas (HGMs) defined by WHO grade III and IV exhibit aggressive biological behavior characterized by high cellular proliferation, nuclear atypia, and increased mitotic activity, collectively resulting in markedly shorter survival than Low-Grade Gliomas (LGMs). It is important for the accurate diagnosis and grading of gliomas for clinical management since grading can have direct consequences for surgical planning, selection of chemotherapy, radiotherapy planning, and prognostication. A brief overview pipeline of the system is shown in Fig. 1.

\begin{figure}[htbp]
    \centering
    \includegraphics[width=0.48\textwidth]{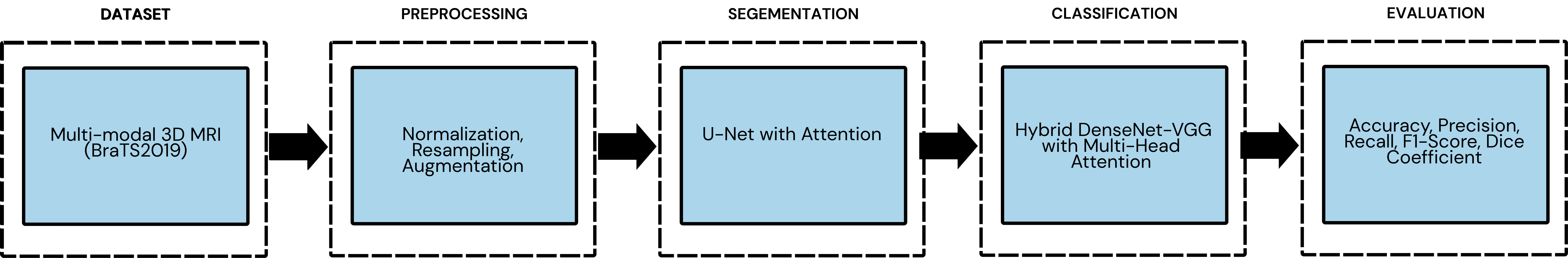}  % path to your image file
    \caption{Research Methodology Process Diagram}
    \label{fig:block_diagram}
\end{figure}

The purpose of this research is to devise a trustworthy glioma detection and grading system that reduces manual errors and time for inspection, while still optimizing for quick clinical outcomes. High dimensional 3D MRI data are efficiently employed through preprocessing methods such as normalization, resampling, and data augmentation. Established metrics are used to evaluate framework performance: Dice coefficient and mean Intersection over Union (IoU) for segmentation performance, and accuracy, precision, recall, and F1-score for classification accuracy. These metrics ultimately enable a higher level of assessment of the robustness and clinical relevance of the framework.

Recent developments in deep learning and machine learning have transformed medical image analysis, which has provided the opportunity to develop advanced automated diagnostic systems that can recognize patterns, extract features, or develop predictive models based on high-dimensional imaging data. Various deep learning architecture approaches have been proposed for glioma analysis: Two-dimensional Convolutional Neural Networks (2D CNN) with VGG19 architecture reached 91.38\% accuracy; three-dimensional Multi-scale CNN approaches achieved 90.64\% accuracy; 3D-based model techniques advanced to 95.31\% accuracy; and three-dimensional Multi-Attention CNN methods achieved the best performance at 95.86\% accuracy. However, none of these prior approaches fully explored the benefits of complementary architectural strengths or leveraged multimodal MRI data with a sophisticated attention methodology approach to provide additional adaptive recalibration of features.

The U-Net model was originally presented in 2015 for the purpose of biomedeical image segmentation, and has proven to be an architecturally very successful model in medical imaging use cases due to its encoder-decoder architecture with skip connections that allows spatial detail to be identified for localization while simultaneously capturing multi-scale contextual features. DenseNet, a year later in 2017, fundamentally altered the design of convolutional neural networks by developing a scheme of dense connectivity for feature reuse and improved gradient propagation enabling the training of very deep networks with fewer overall parameters. VGG networks, in contrast, have a significant depth while maintaining comparatively smaller receptive fields which allow them to hierarchically create increasingly more abstract features from many successive transformations. Multi-head attention, now found in the transformer family of architectures, allowed the model to attend to multiple aspects of multiple different features and multiple different spatial regions, which could be adapted to the model focusing on tumor features that are determined to be clinically relevant.

This study introduces a new hybrid deep learning model which integrates tumor segmentation using a 3D U-Net with a dual-branch hybrid DenseNet-VGG classification model with multiple heads and spatial-channel attention. The framework integrates multi-modal 3D MRI data within a pipeline model: the 3D U-Net segmented the anatomy from volumetric MRI data while ensuring the spatial context was maintained and output the segmented anatomy to the hybrid classification model, where complementary DenseNet and VGG branches, enhanced by attention mechanisms, learn to extract deferential tumor features and pay attention to proximal anatomy associated clinically. This combination of structures exploits complementarity, and integrates an attention model, is expected to achieve better performance.

Three-dimensional volumetric analysis diverges from the more conventional two-dimensional slice-by-slice study, as features of the tumor such as the morphology, spatial relationships, and volumetric traits are fully captured in 3D processing, whereas those elements may be absent or distorted in 2D representation. Moreover, This formulated newer 3D visualization work allows a more complete characterization of tumors including volume methods, shape descriptors, and distances between tumor subcomponents; which, although less measurable, are still relevant for surgical procedure deliverables, and assessment of clinical treatment. BraTS2019 is a publicly available database consisting of 335 annotated multimodal 3D MRI scans with 5 expert annotations for each patient (259 high grade glioma, and 76 low grade glioma).

As applied, the frameworks performance is measured quantitively via metrics accepted in scientific and medical imaging research communities. For example, segment performance is measured via the Dice coefficient and mean Intersection over Union (mIoU) results. Classification accuracy is represented with additional performance metrics such as accuracy, precision, recall, specificity, F1-score, etc. The proposed work achieves an overall Dice coefficient higher than 98\% for segmentation accuracy and over 99\% classification accuracy for glioma grade while significantly outperforming stated of the art segmentation methods. This work reinforces the idea of using both deep learning architecture combinations with an advanced attention mechanism to analyze medical images.

This research adds to the literature through: (1) a novel hybrid structure of U-Net, DenseNet, and VGG that incorporates a variety of attention mechanisms; (2) extensive analyses of 3D MRI, using an extensive collection of multivariate volumetric data; (3) best in class performance that demonstrates significant advantages over previously published results; (4) thorough evaluations that use established metrics common among medical imaging; and (5) thorough, potentially reproducible and translatable approaches. The framework acts on essential clinical gaps by enabling a fast, automated, objective glioma detection and grading route that helps to decrease diagnostic error rates, expediting and improving clinical workflow, and enhancing outcomes with quick accurate treatment planning.

\section{Literature Review}

Among brain tumours, gliomas are some of the most deadly tumours; thus, early-emergent detection and proper grading are significant factors for their treatment. The previous literature discusses using deep learning approaches to improve the segmentation and classification of gliomas derived from MRI.

Tripathi and Bag's \cite{tripathi2022} study introduced an attention-guided CNN model that utilized a multi-task learning approach with spatial and channel attention blocks to segment gliomas and classify them as low/high grade, 1p/19q, and IDH status. Their model segmented the tumor and used that segmented tumour for the classification to achieve better accuracy than existing methods or pipelines, yet the glioma grading was not based on the robust increase from the 1p/19q status.

Some studies propose deep learning pipelines that will apply preprocessing techniques prior to segmentation techniques. As an example, Dang et al. \cite{dang2022} segmented the glioma using U-Net and classified the glioma using either VGG or GoogleNet, achieving over 97.44\% accuracy. Vinaya and Mara \cite{vinaya2022} applied transfer learning using EfficientNet for classification and SegNet for segmentation, achieving 97\% accuracy or Dice coefficient of 0.81, but only for low-grade gliomas. Naser and Deen \cite{naser2020} combination of U-Net segmentation and transfer learning using VGG16 for the lower-grade gliomas achieved 0.84 mean DSC and 0.89 image level accuracy, although fine-tuning was required for optimal accuracy.

Multiple articles examine multimodal 3D MRI segmentation. Trivedi et al. \cite{trivedi2022} proposed NDNN-based U-Net for volumetric segmentation, with Dice scores of up to 90.02\%. Ilyas et al. \cite{ilyas2022} demonstrated Hybrid-DANet, a network based on multi-dilated attention and hybrid weight alignment building upon the advantages of segmented attention as a method to improve segmentation accuracy at higher computational cost than previously developed models. Hapsari et al. proposed a DNN framework with minimal texture features on feature extraction (GLCM, GLRLM, GLSZM) to classify and evaluate tumours compared to SVM, though careful training of the DNN state depends on adaptation. Semi-supervised multi-task learning approaches to genomic subtyping that predicted genomic tumor markers such as IDH, MGMT, and 1p/19q were proposed by Tupe-Waghmare et al. \cite{tupe2021}, achieving approximately 82.35\% accuracy compared to moderate Dice scores.

Previous studies have examined deep learning based segmentation and classification with unique preprocessing and feature techniques. \"{O}zkaya and S\u{a}\u{g}iro\u{g}lu \cite{ozkaya2023} identified an improvement of 15\% Dice score employing adaptive histogram thresholding for tumour segmentation. Ghassemi et al. proposed a GAN based methodology for classification pretraining. Oinar and Yildirim modified ResNet-50 for binary detection of glioma. Several studies address limited datasets, binary classification, coarse spatial grading of tumors, findings were developed by Mohsen et al., Sajid et al., Irmak, and Ayadi et al.

In general, deep learning methodologies for glioma detection and grading have progressed considerably due to attention mechanisms, transfer learning, and hybrid models. However, the challenges of small datasets, high computational cost, and limited use of multimodal and genetic data remain. Credibly, the hybrid 3D CNN U-Net framework for accurate segmentation and multi-level classification, represents an exciting future direction of possible work.

\subsection{Research Objectives}

To develop a glioma detection model for Segmentation and classification of 3D MRI.
Reduce inspection time \& manual error in diagnosis using the AI Attention mechanism.
Optimize the system for real-time usage and increase accuracy.

\subsection{Research Challenges}

The project major challenge is designing glioma detector from 3D MRI's, a deep learning model particularly made for glioma segmentation and grading.
The working of the proposed system will be evaluated with the helping metrics such as segmentation accuracy, Dice similarity coefficient, grading accuracy etc., which is supposed to be more.
Complicated dataset of 3D MRI scans of glioma patients will be collected and used for training plus validation purposes.

\section{Proposed Architecture}

The primary research strategy implemented is to classify and compare the glioma and decide its grade for medical treatment. The training of the model-3D MRI images is necessary and fundamental. The data set is collected from an open source, and we use freely accessible data sets. Proposed architecture is shown in Fig. 2.

\begin{figure}[htbp]
\centerline{\includegraphics[width=0.48\textwidth]{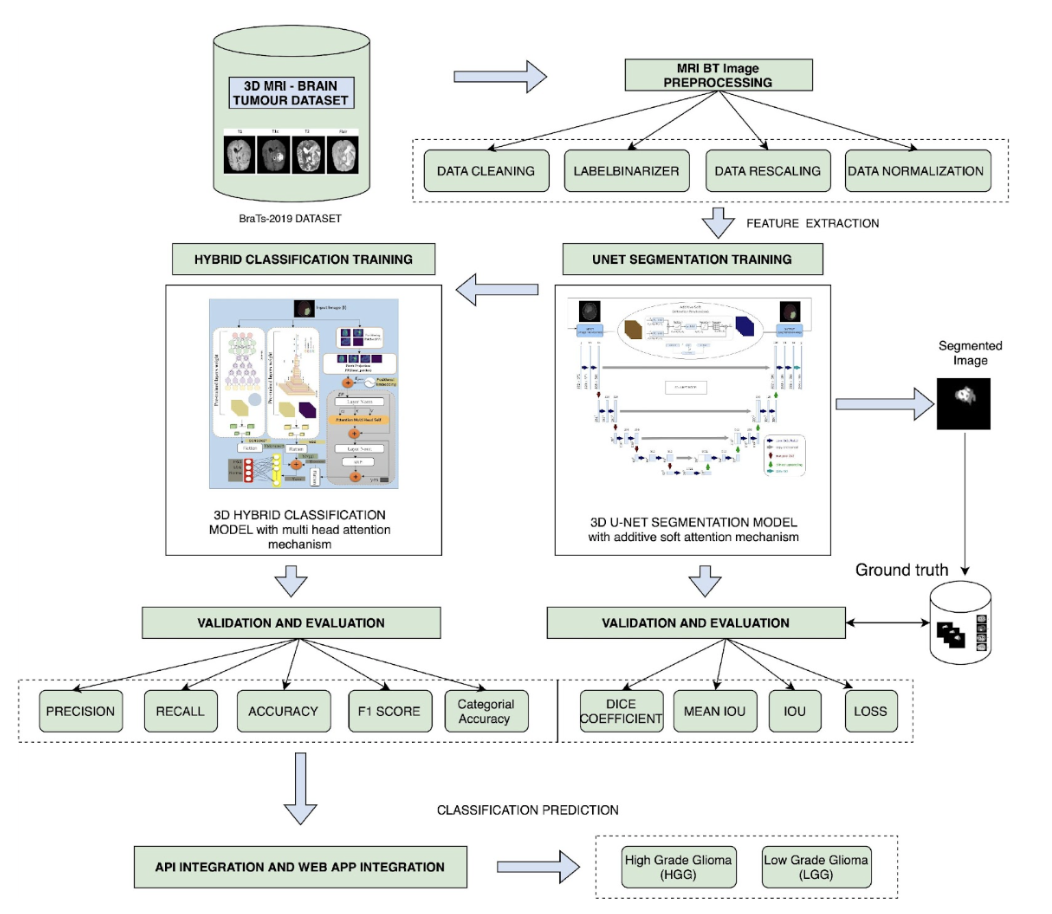}}
\caption{Architecture of System.}
\label{architecture}
\end{figure}

BRATS2019 dataset. All images in data set is divided into training \& testing data. Glioma in the high defined 3D MRI image is spotted using the U-net deep learning model. Then, we downsampling the images for clear identification of encoder path. Similarly, upsampling them highlights the decoder path. This process leads to pixel wise classification making sure that image input and output sizes are consistent. The encoder is incharge of ``WHAT'' is displayed within the image and the decoder is incharge of ``WHERE'' is glioma area pinpointed. After collecting the dataset, images undergo preprocessing and segmentation that too into different classes. Besides training, gliomas region is highlighted and utilised for classification. Finally we use different metrics to compare performance of the model. The flow chart in Figure 1 outlines the proposed system working from beginning to conclusion.

\begin{algorithm}[!ht]
\caption{Hybrid CNN--Attention Model for Glioma Classification}
\label{alg:hybrid_cnn_attention}
\begin{algorithmic}[1]
\REQUIRE Segmented 3D brain MRI volumes X, ground truth labels Y
\ENSURE Resulting trained model M

\STATE Input: Preprocessed and segmented MRI scans of glioma regions.
\STATE Base Model Initialization: Load weights of U-Net backbone for feature extraction.
\STATE Dense Block Construction: Each layer receives output from all previous layers concatenated together:
\[
F_l = H_l([F_0, F_1, \ldots, F_{l-1}])
\]
\STATE Batch Normalization: Normalization of activations for training stability.
\[
Y = \frac{X - \mu}{\sqrt{\sigma^2 + \varepsilon}} \gamma + \beta
\]
where $\mu$ and $\sigma$ is the mean and variance across the batch, respectively, and $\gamma$, $\beta$ are learnable.
\STATE Pooling and Transition: Downsample feature map size while retaining important features in the representation:
\[
N_{out} = \left(\frac{N_{in} + 2p - k}{s}\right) + 1
\]
\STATE Attention Module: The attention on channel and spatial dimensions is generated with:
\[
\text{Attention}_h = \text{softmax}\left(\frac{Q_h K_h^T}{\sqrt{d_k}}\right)V_h
\]
\STATE Global Average Pooling (3D): Aggregate the spatial information across depth, height, and width as:
\[
\text{GAP}_z = \frac{1}{H \times W} \sum_{i=1}^{H} \sum_{j=1}^{W} X_{ijz}
\]
\STATE Fully Connected Layer: Combine the high-level learned features for the classification task:
\[
\text{Output} = \text{Dense}(n_{\text{classes}}, \text{activation=softmax})(\text{GAP3D}(x))
\]
\STATE Loss Function: Categorical cross-entropy is calculated as:
\[
\mathcal{L}(y, \hat{y}) = - \sum_i y_i \log(\hat{y}_i)
\]
\STATE Optimization: Parameters are updated using Adam as:
\[
\theta_{t+1} = \theta_t - \eta \frac{m_t}{\sqrt{v_t} + \epsilon}
\]
\STATE \textbf{return} Trained hybrid CNN--attention model M*
\end{algorithmic}
\end{algorithm}

\section{Methodology}

\subsection{Research Method Overview}

This study proposes a comprehensive hybrid deep learning framework specifically for automated glioma segmentation and classification of 3D MRI volumes. The proposed approach integrates a tailored segmentation model with a sophisticated hybrid classification model, resulting in a streamlined end-to-end workflow for processing raw medical imaging data into clinically relevant tumor grade predictions. The framework processes high dimensional 3D MRI data using a five-step sequential process which consists of: (1) acquiring and fully characterizing the dataset, (2) preprocessing and normalizing the data consistently, (3) segmenting the tumor using a 3D U-Net model with added attention, (4) classifying the MRI through a features-based approach using a hybrid DenseNet-VGG model with multi-head and spatial-channel attention, and (5) quantitatively evaluate the models using established medical imaging evaluation metrics.

In terms of the research approach, a supervised learning approach was used to train both the segmentation and classification models on annotated data with ground truth labels. This approach was motivated by four objectives: (i) minimizing error and time taken by the radiologist during MRI assessment, (ii) taking advantage of complementary capabilities of multiple deep learning architectures through a hybridized model, (iii) using attention-enhanced models to increase model learning on clinically accepted tumor features, and (iv) establishing state-of-the-art performance on benchmark datasets with a broader audience for clinical utility. Methodology of the system is shown in Fig. 3.

\begin{figure}[htbp]
    \centering
    \includegraphics[width=0.48\textwidth]{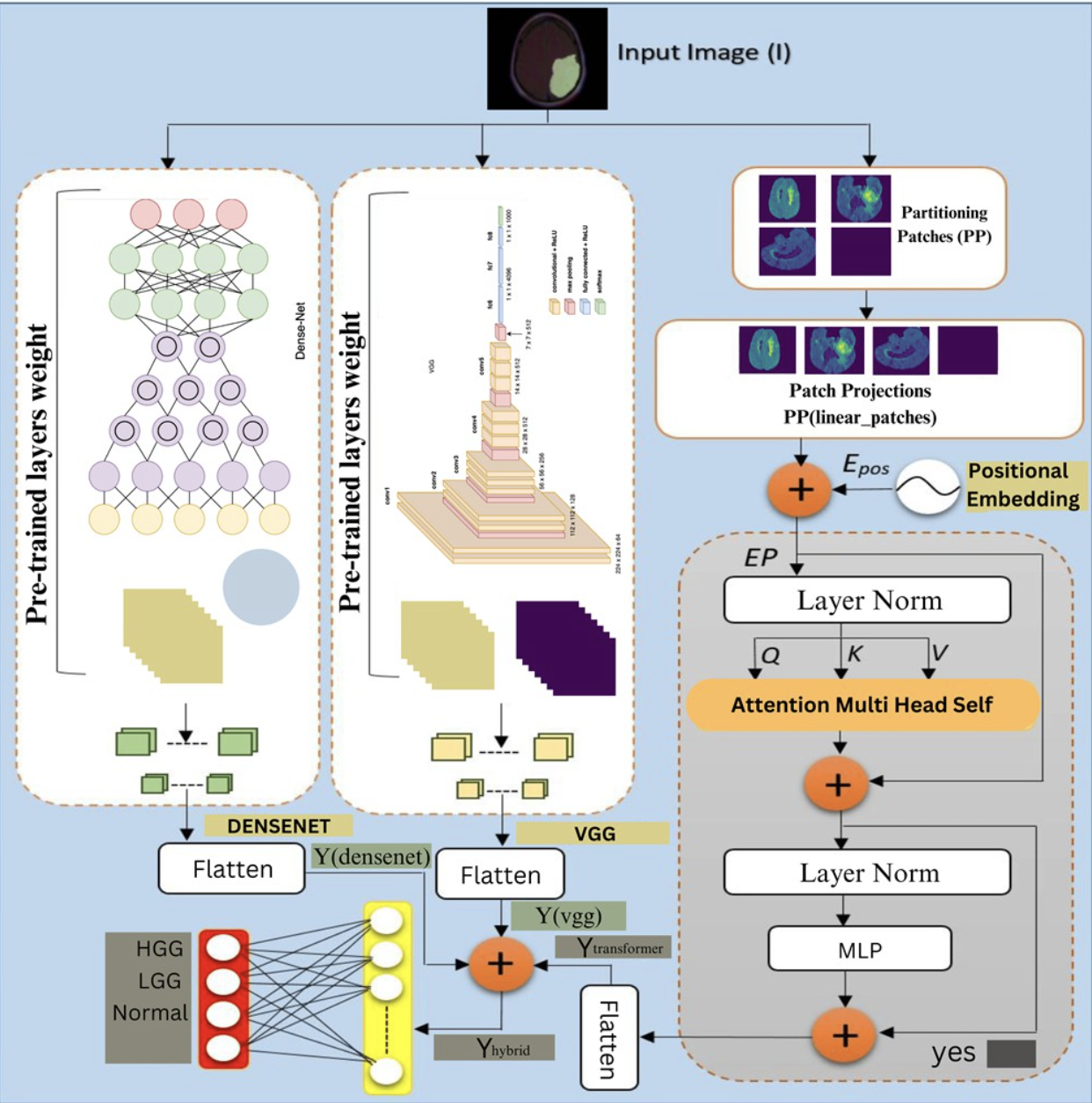}  % path to your image file
    \caption{Research Methodology Process Diagram}
    \label{fig:method}
\end{figure}

\subsection{Dataset Description and Preparation}

\subsubsection{BraTS2019 Dataset Characteristics}

The BraTS2019 dataset is a widely used public resource featuring 335 multi-modal 3D MRI scans from various clinical centers. Each scan includes five imaging types: FLAIR, T1-weighted, T1-weighted with gadolinium contrast (T1ce), T2-weighted, and a manual tumor segmentation mask. Of these 335 scans, 259 are from patients with aggressive High-Grade Gliomas (HGG), and 76 are from patients with less aggressive Low-Grade Gliomas (LGG), representing the typical clinical ratio.

Each MRI modality highlights different aspects of the tumor. FLAIR highlights swelling and abnormal tissues by suppressing fluid signal, while T1 provides detailed anatomical structure. T1ce reflects tumor regions with active blood-brain barrier disruption through contrast enhancement. T2 shows non-enhancing and cystic tumor areas. The segmentation masks, expertly annotated by radiologists, distinguish tumor presence versus healthy brain tissue.
Fig. 4 shows a sample of the Dataset:

\begin{figure}[htbp]
    \centering
    \includegraphics[width=0.48\textwidth]{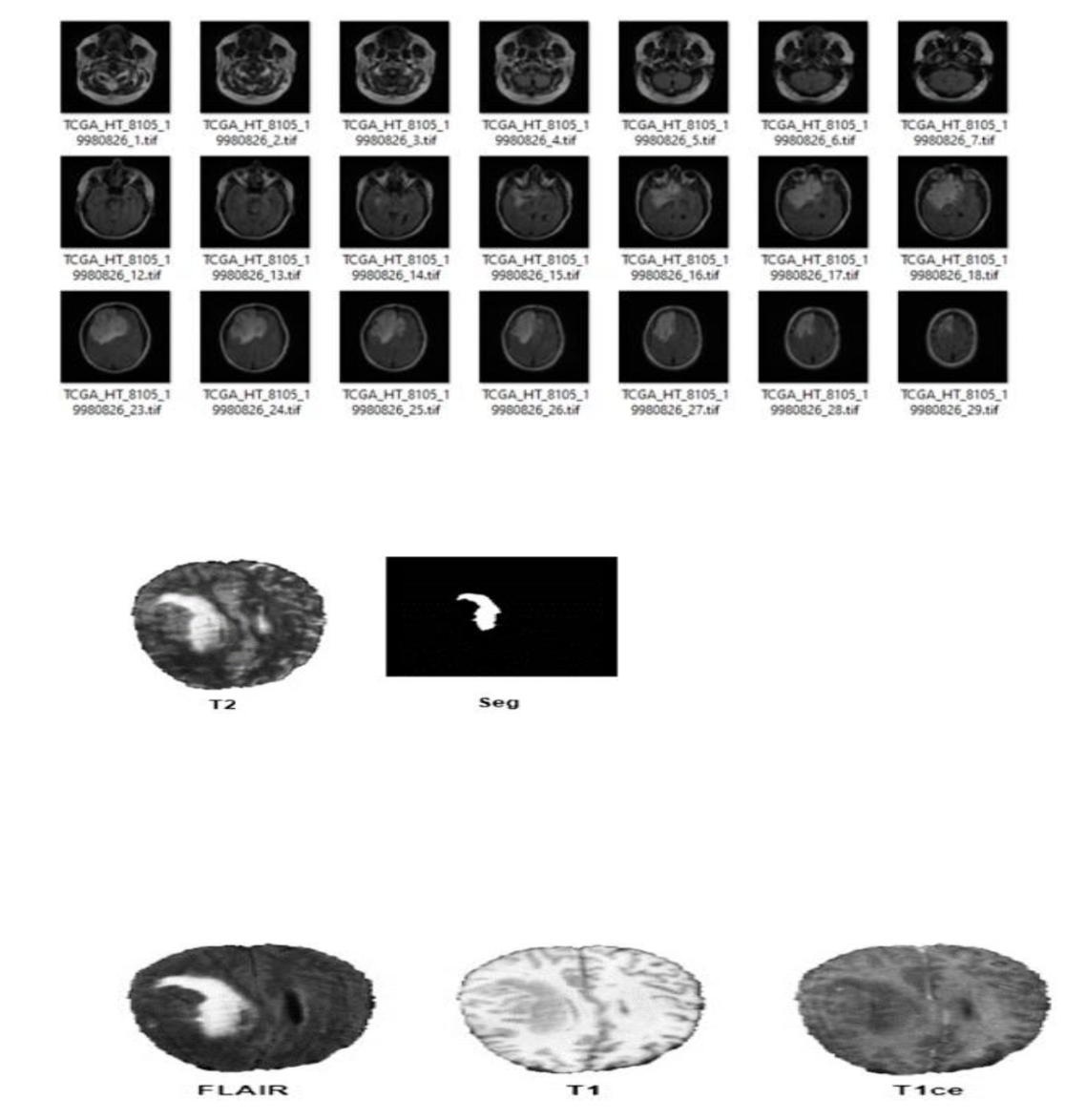}  % path to your image file
    \caption{Dataset}
    \label{fig:dataset}
\end{figure}

\subsubsection{Data Preprocessing Strategy}

The raw MRI volumes in BraTS2019 have 155 consecutive axial slices, each with spatial dimensions of $240 \times 240$ pixels and isotropic voxel spacing of $1\,\text{mm}^3$. To enable viewing and utilizing diagnostic material on a GPU with limited graphics memory, all volumes are preprocessed to standardize formatting.

\textbf{Volume Resampling:} All 3D MRI volumes are resampled to $128 \times 128 \times 1$ voxels using trilinear interpolation, rather than using all 155 slices. In preprocessing, the pipeline selects the central 64 consecutive slices in which the maximum tumor burden appears by automated analysis of ground truth masks to produce a smaller reformatted volumetric image that saves memory while also including the slices most pertinent to analyzing the tumor. This oversampled volumetric image reduces memory requirements by nearly 58\% over processing a full volume image.

\textbf{Intensity Normalization:} Raw magnetic resonance imaging (MRI) intensities can be quite variable across differing scanning protocols, scanner vendors, and acquisition parameters, which creates distribution shift that confounds convergence for neural networks to train. A normalization of z-scores is applied on a per-subject, per-modality basis, for example,

\begin{equation}
I_{\text{norm}}(x,y,z) = \frac{I_{\text{raw}}(x,y,z) - \mu}{\sigma + \epsilon}
\end{equation}

where $I_{\text{raw}}(x,y,z)$ denotes raw intensity at voxel location $(x,y,z)$, $\mu$ and $\sigma$ represent mean and standard deviation computed exclusively within the brain mask (excluding background), and $\epsilon = 1 \times 10^{-6}$ avoids numerical instability. This normalization centers each modality distribution to a zero mean and unit variance, allowing for robust cross-subject and cross-scanner comparisons while maintaining enough dynamic range in the distribution for feature learning.

\textbf{Label Binarization:} While BraTS2019 does have fine-grained tumor subregion annotations (enhancing core, necrotic center, edema), the study will only conduct a binary classification (presence vs. absence of tumor). As a result, all ground truth labels will be binarized: 

\begin{equation}
L_{\text{binary}}(x,y,z) = \begin{cases} 
1 & \text{if } L_{\text{original}}(x,y,z) > 0 \\
0 & \text{otherwise}
\end{cases}
\end{equation}

\textbf{Data Augmentation:} To enhance model robustness on limited training data (335 total subjects), comprehensive augmentation is applied exclusively to training data during each epoch:

\begin{itemize}
\item \textbf{Spatial Augmentation}: Random 3D rotations ($\pm 15°$) and translations ($\pm 10$ voxels) account for natural anatomical variations and scanner positioning differences
\item \textbf{Elastic Deformation}: B-spline-based spatial deformations simulate morphological variations in tumor presentation
\item \textbf{Intensity Augmentation}: Random multiplicative scaling (0.95--1.05) and additive Gaussian noise ($\sigma = 0.02$) simulate scanner variability
\item \textbf{Contrast Stretching}: Random contrast enhancement simulates different reconstruction algorithms
\end{itemize}

The entirety of the dataset is divided into training (75\%, $n=251$ patients) and validation (25\%, $n=84$ patient) sets via stratified random sampling at the level of the patient. Notably, all the MRI slices associated with a single patient will reside either in the training or validation set ensuring that there is no information leakage that would overly inflate performance estimates. Stratification maintains the equivalent class distributions between partitions.

\subsection{Tumor Segmentation Architecture}

\subsubsection{3D U-Net Foundation}
The segmentation of tumors uses a 3D U-Net model with two components of the U-Net design. There are 23 convolutions sequenced as an encoder-decoder network. The encoder module gradually reduces the size of 3D feature representations in a pyramidal fashion over 5 levels. Each level consists of 2 sequential convolutional layers with kernels of size 3x3x3. Each convolutional layer is followed by a ReLU and a max-pooling layer with a stride of 2. The downward spatial reduction of 3D feature representations, for example, describes information multi scaly-dependent context which is key to or serves as an indication of possible tumor boundaries. The decoder has a corresponding structure to the encoder part of the U-Net, and the decoder will progressively up-sample next feature maps through transposed convolutions with a stride of 2.

The inclusion of skip connections is another architectural characteristic of U-Net features that takes the encoder features (down-sampled) and concatenates them prior to up-sampling with the corresponding features in the decoding branch of the U-Net model. Skip connections will preserve spatial information lost during the down-sampling process and will allow for voxel-level tumor boundary detection. This building block of design is important for medical imaging and medical imaging segmentation in practice, especially when boundaries delineating the tumor are clinically relevant to attain during diagnosis and treatment strategies. A prediction of test image of the system is illustrated in Fig. 5.

\begin{figure}[htbp]
    \centering
    \includegraphics[width=0.48\textwidth]{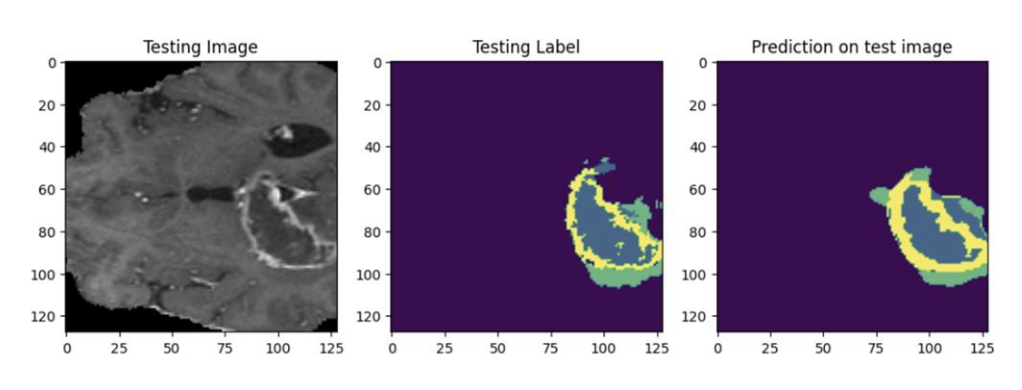}  % path to your image file
    \caption{U-Net Model.}
    \label{fig:unet}
\end{figure}

\subsubsection{Soft Additive Attention Integration}
The Standard U-Net operates on all spatial locations with equal importance. To allocate network capacity for tumor-relevant areas while suppressing background information, soft additive attention is added to the segmentation pipeline is shown in Fig. 6:

\begin{figure}[htbp]
    \centering
    \includegraphics[width=0.48\textwidth]{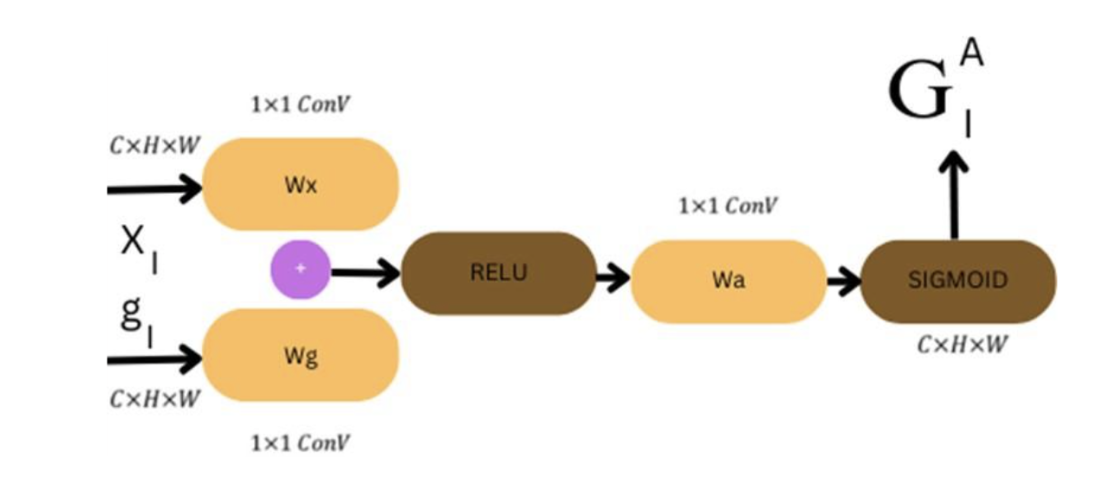}  % path to your image file
    \caption{Attention Mechanism.}
    \label{fig:attention}
\end{figure}

\begin{equation}
A_{\text{soft}}(x) = \sigma(W \cdot f(x))
\end{equation}

where $\sigma(\cdot)$ denotes sigmoid activation constraining outputs to [0,1], $W$ is a learnable weight matrix, and $f(x)$ represents feature activations at spatial location $x$. The computed attention mask $A_{\text{soft}}$ exhibits higher values for tumor regions and lower values for background, enabling adaptive feature recalibration:

\begin{equation}
f_{\text{attended}}(x) = A_{\text{soft}}(x) \odot f(x)
\end{equation}

where $\odot$ denotes element-wise multiplication. This mechanism effectively reweights spatial locations, directing network focus toward clinically informative tumor regions while suppressing irrelevant background noise.

\subsubsection{Segmentation Training and Loss}

The segmentation network output applies sigmoid activation, generating pixel-wise tumor probability estimates $p(x,y,z) \in [0,1]$:

\begin{equation}
\text{Sigmoid}(Q) = \frac{1}{1 + \exp(-Q)}
\end{equation}

Binary cross-entropy loss optimizes the segmentation network:

\begin{equation}
\mathcal{L}_{\text{BCE}} = -\frac{1}{N} \sum_{i=1}^{N} \left[ y_i \log(p_i) + (1-y_i)\log(1-p_i) \right]
\end{equation}

where $N$ denotes total voxel count, $y_i \in \{0,1\}$ is ground truth label, and $p_i$ is model prediction. This logarithmic formulation applies higher penalty to confident misclassifications, naturally addressing class imbalance (background voxels vastly outnumber tumor voxels).

The segmentation network is trained using the Adam optimizer with learning rate $\alpha = 0.001$, batch size 16, and up to 100 epochs. Early stopping based on validation loss prevents overfitting.

\subsubsection{Segmentation Evaluation Metrics}

Segmentation quality is quantitatively assessed using complementary metrics:

\textbf{Dice Coefficient:} Measures spatial overlap between predicted segmentation $P$ and ground truth $Q$:

\begin{equation}
\text{Dice} = \frac{2|P \cap Q|}{|P| + |Q|}
\end{equation}

Dice ranges from 0 (no overlap) to 1 (perfect agreement). In clinical applications, Dice $\geq 0.90$ represents excellent segmentation quality. This metric is robust to class imbalance common in medical imaging.

\textbf{Mean Intersection over Union (mIoU):} Computes the intersection-over-union metric, emphasizing prediction precision:

\begin{equation}
\text{mIoU} = \frac{|P \cap Q|}{|P \cup Q|} = \frac{|P \cap Q|}{|P| + |Q| - |P \cap Q|}
\end{equation}

mIoU is particularly sensitive to false positive predictions, providing stricter evaluation than Dice coefficient and better reflecting clinical utility where false positives can trigger unnecessary clinical interventions.

\subsection{Hybrid Classification Architecture}

\subsubsection{Architectural Design Rationale}
After tumor segmentation, the segmented regions of the tumor are classified using a novel hybrid network based on DenseNet and VGG architectures. This hybrid architecture exploits their architectural strengths: DenseNet's efficient feature reuse and gradient propagation, and VGG's depth to extract hierarchical features at a finer scale.

\textbf{DenseNet Component:} DenseNet implements dense connectivity where each layer receives concatenated outputs from all preceding layers:

\begin{equation}
X_l = H_l([X_0, X_1, \ldots, X_{l-1}])
\end{equation}

where $[X_0, X_1, \ldots, X_{l-1}]$ denotes concatenation of outputs from layers 0 through $l-1$, and $H_l$ Denotes the composition function (batch normalization $\rightarrow$ ReLU $\rightarrow$ convolution). The advantages of a denser connectivity network is: (i) that it enables training of very deep networks by improving gradient flow through the many short paths; (ii) reusing features across layers provides parameter efficiency; (iii) reusing features provides implicit regularization which can dampen overfitting. Accordingly, DenseNet is able to improve features and distinguish among glioma subtypes.

\textbf{VGG Component:} VGG networks employ deeper architectures with small $3 \times 3$ receptive fields:

\begin{equation}
f_{\text{hierarchical}}(x) = f_L(f_{L-1}(\cdots f_1(x) \cdots))
\end{equation}

where $L$ refers to network depth. VGG's progressive increase in network depth allows the extraction of increasingly abstract hierarchical features, with the early layers capturing low-level features (edges, textures) and the deeper layers capturing higher level semantic features (i.e., shape, patterns of intensity). Although VGG requires more parameters than DenseNet, the depth of VGG helps to discriminate subtle features that are crucial for distinguishing glioma grades.

\textbf{Hybrid Integration:} The designed architecture uses DenseNet and VGG as two separate, distinct branches that process segmented features of the tumors. Each branch extracts valuable complementary feature hierarchies that represent different components of the tumor. After deriving branch outputs, we concatenate the outputs and input them into multiple layers to promote fusion of these complementary representations before making the final classification.

\subsubsection{Multi-Head Attention Mechanism}

Multi-head attention enables simultaneous attention to multiple feature aspects through parallel computation across diverse representation subspaces:

\begin{equation}
\text{MultiHeadAttention}(Q, K, V) = \text{Concat}(\text{head}_1, \ldots, \text{head}_h) W^O
\end{equation}

Each attention head computes scaled dot-product attention:

\begin{IEEEeqnarray}{rCl}
\text{head}_i 
& = & \text{Attention}(Q W_i^Q, K W_i^K, V W_i^V) \nonumber \\
& = & \text{softmax}\left(\frac{Q W_i^Q (K W_i^K)^T}{\sqrt{d_k}}\right) V W_i^V
\end{IEEEeqnarray}

where $Q$, $K$, $V$ denote query, key, and value feature map projections; $W_i^Q, W_i^K, W_i^V$ are learnable projection matrices for head $i$; $d_k$ is key dimension; scaling by $\sqrt{d_k}$ stabilizes gradient magnitudes; and $W^O$ linearly combines multi-head outputs.

Since multiple attention heads are working in parallel, different heads focus their attention on different aspects of the features: some heads focus on tumor morphology, some heads focus on the distributions of intensity, and some focus on spatial configuration. This aspect of diversity improves representational capacity so that the network can also learn to attend to multiple tumor characteristics that may have clinical significances at once.

\subsubsection{Spatial and Channel Attention}

Complementing multi-head attention, spatial and channel attention mechanisms provide additional feature recalibration:

\textbf{Spatial Attention:} Generates attention masks emphasizing clinically important spatial regions:

\begin{equation}
A_{\text{spatial}}(x,y,z) = \text{sigmoid}(f_{\text{conv}}(\text{MaxPool}(F) + \text{AvgPool}(F)))
\end{equation}

represents 3D feature maps, MaxPool and AvgPool summary max and average activations across spatial dimensions and $f_{\text{conv}}$ applies learnable convolution. Higher attention values indicate more importance for classification.

\textbf{Channel Attention:} Recalibrates feature importance across channels:

\begin{equation}
A_{\text{channel}} = \text{sigmoid}(W_2 \text{ReLU}(W_1(\text{AvgPool}(F) + \text{MaxPool}(F))))
\end{equation}

where $W_1, W_2$ are learnable fully-connected layer weights. Channel attention enables selective emphasis on feature channels maximally informative for glioma grading.

Combined spatial-channel attention applies multiplicative interaction:

\begin{equation}
F_{\text{refined}} = (A_{\text{spatial}} \odot A_{\text{channel}}) \odot F
\end{equation}

This two-attention approach allows full calibration of the features along both spatial and channel dimensions to focus on clinical characteristics of the tumor that are relevant. 

\subsubsection{Classification Output and Prediction}

Following feature refinement, global average pooling aggregates spatial information:

\begin{equation}
\text{GAP}(F) = \frac{1}{H \times W \times D} \sum_{i=1}^{H} \sum_{j=1}^{W} \sum_{k=1}^{D} F(i,j,k)
\end{equation}

where $H, W, D$ denote spatial dimensions of feature maps. This operation converts spatial feature maps to vector representations, providing position-invariant features essential for classification.

The final classification layer applies softmax activation:

\begin{equation}
\text{Output} = \text{softmax}\left(W_c \cdot \text{GAP}(F) + b_c\right) = \frac{\exp(z_i)}{\sum_j \exp(z_j)}
\end{equation}

where $W_c, b_c$ are learnable classification weights and biases. Softmax normalization ensures output values represent valid probability distributions. The network outputs two probabilities: $P(\text{HGG})$ and $P(\text{LGG})$, with the higher probability determining final grade prediction.

\subsubsection{Classification Training}

Categorical cross-entropy loss optimizes the classification network:

\begin{equation}
\mathcal{L}_{\text{CE}} = -\sum_{i=1}^{C} y_i \log(\hat{y}_i)
\end{equation}

where $C = 2$ (binary classification: HGG or LGG), $y_i$ denotes one-hot encoded ground truth (1 for true class, 0 otherwise), and $\hat{y}_i$ is softmax-normalized prediction.

The classification network is trained using the Adam optimizer with learning rate $\alpha = 0.0005$, batch size 8, and up to 150 epochs. The lower learning rate compared to segmentation training enables more careful feature learning. Early stopping based on validation accuracy prevents overfitting.

\subsection{Quantitative Evaluation Framework}

\subsubsection{Classification Performance Metrics}

Classification quality is comprehensively assessed using multiple metrics:

\textbf{Accuracy:} Overall fraction of correct predictions:

\begin{equation}
\text{Accuracy} = \frac{\text{TP} + \text{TN}}{\text{TP} + \text{TN} + \text{FP} + \text{FN}}
\end{equation}

\textbf{Precision:} Fraction of positive predictions that are correct:

\begin{equation}
\text{Precision} = \frac{\text{TP}}{\text{TP} + \text{FP}}
\end{equation}

High precision indicates few false alarms (false HGG diagnoses).

\textbf{Recall (Sensitivity):} Fraction of actual positive cases correctly identified:

\begin{equation}
\text{Recall} = \frac{\text{TP}}{\text{TP} + \text{FN}}
\end{equation}

High recall indicates the model detects most actual HGG cases, critical for clinical safety.

\textbf{Specificity:} Fraction of negative cases correctly identified:

\begin{equation}
\text{Specificity} = \frac{\text{TN}}{\text{TN} + \text{FP}}
\end{equation}

\textbf{F1-Score:} Harmonic mean of precision and recall, balancing both metrics:

\begin{equation}
\text{F1} = 2 \cdot \frac{\text{Precision} \times \text{Recall}}{\text{Precision} + \text{Recall}}
\end{equation}

where TP (true positives), TN (true negatives), FP (false positives), and FN (false negatives) are computed from validation predictions versus ground truth labels.

\subsubsection{Segmentation Performance Metrics}

Segmentation quality employs previously defined Dice coefficient and mean Intersection over Union metrics, evaluated on validation tumor masks.

\subsection{Evaluation Metrics}
Model performance is assessed using standard metrics:

\begin{align}
\text{Accuracy} &= \frac{\sum_b (TP_b + TN_b)}{\sum_b (TP_b + TN_b + FP_b + FN_b)} \\
\text{Precision} &= \frac{\sum_b TP_b}{\sum_b (TP_b + FP_b)} \\
\text{Recall} &= \frac{\sum_b TP_b}{\sum_b (TP_b + FN_b)} \\
F_1 &= 2 \cdot \frac{\text{Precision} \cdot \text{Recall}}{\text{Precision} + \text{Recall}} \\
\text{Specificity} &= \frac{TN}{TN + FP}
\end{align}

The Dice coefficient and mIoU are used as metrics to evaluate segmentation and the hybrid model shows exceptional HGG versus LGG classification and accurate glioma region localization from preprocessed 3D MRI data.

\section{Results}

We evaluated the proposed hybrid classification model by utilizing loss, accuracy, precision, and F1-score metrics over training time. The precision graph examines how accurately the model was predicting the metric over all epochs. The accuracy variable would increase over training time, indicating that the model's abilities to predict gliomas increased over training time.

The loss graph measures the predicted value vs actual value. A decreasing loss accounts for the model's ability to become more converged to sophisticated performance. The accuracy should be consistently high for reliable classification of glioma images. The proposed hybrid classification model accurately provides the format to perform the task of reliable localization and classification of whether glioma is present on medical images once we achieve a classification accuracy of 99.99\% percent.

\begin{figure}[htbp]
\centerline{\includegraphics[width=0.48\textwidth]{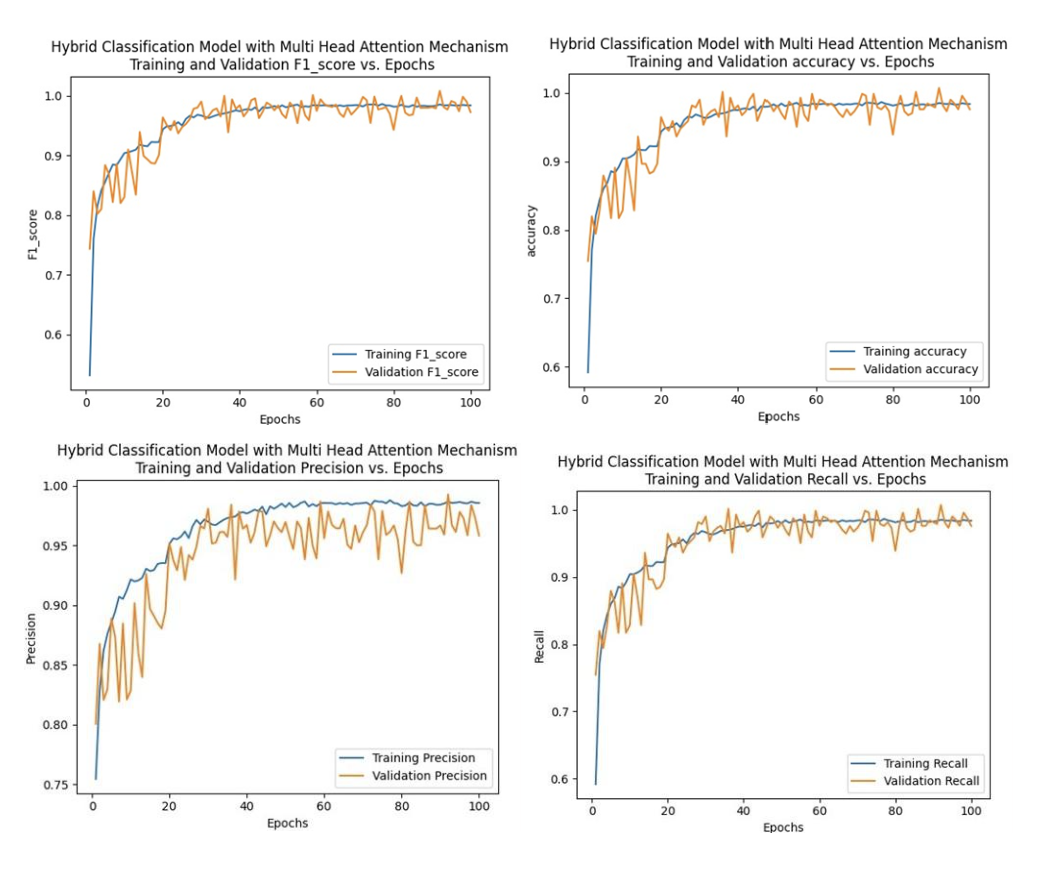}}
\caption{Categorical Accuracy vs Epoch Graph for Multi-Head Attention Mechanism.}
\label{fig:epoch_accuracy}
\end{figure}

\begin{figure}[htbp]
\centerline{\includegraphics[width=0.48\textwidth]{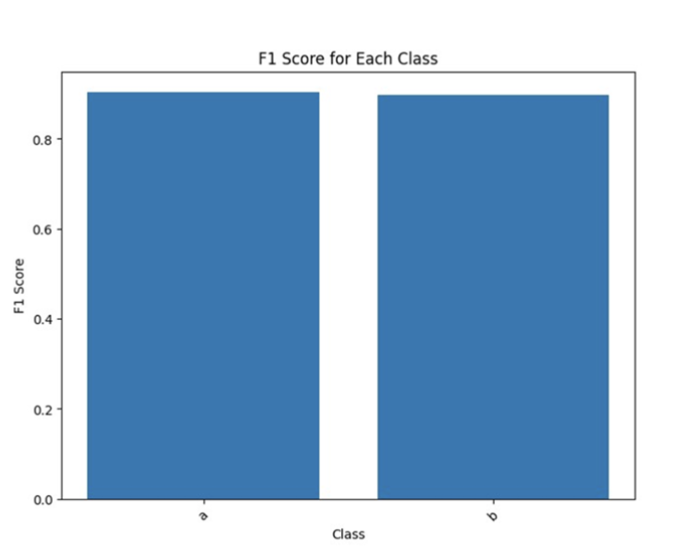}}
\caption{F1 Score of Each Class.}
\label{fig:class_f1}
\end{figure}

\subsection{Performance Evaluation}
Commonly used as performance metrics for classification models, confusion metrics, or confusion matrices, were employed in this study. We employed true positive (TP), true negative (TN), false positive (FP), and false negative (FN) counts to evaluate performance in a hybrid classification model with DenseNet and VGG classification components. These classifications allowed us to assess the model's performance in classifying if a glioma case was predicted correctly (TP), accurately classifying a case with no glioma (TN), inaccuracies by the model classifying a case with no glioma as glioma (FP), and incorrect classifications if a case with glioma was misclassified as a case with no glioma (FN). Model Performance is shown in Fig. 7.

\subsubsection{Sensitivity and Specificity}

Sensitivity is defined as:

\begin{equation}
\text{Sensitivity} = \frac{SP}{SP + EN}
\end{equation}

where $P_1$ represents the predicted glioma region and $T_1$ is the ground truth region of glioma.  

Specificity is defined as:

\begin{equation}
\text{Specificity} = \frac{SN}{SN + EP}
\end{equation}

where $SN$ represents true negatives, $T_0$ is the non-glioma region of the ground truth, and $P_0$ is the predicted non-glioma region.

\subsubsection{Dice Coefficient}

The Dice coefficient, commonly used for segmentation evaluation, is defined as:

\begin{equation}
\text{Dice coefficient} = \frac{2 \times |P \cap Q|}{|P| + |Q|}
\end{equation}

\subsubsection{Accuracy, Precision, and F1-Score}

Accuracy measures the overall correctness of the classification:

\begin{equation}
\text{Accuracy} = \frac{\sum_{b=1}^{B} (TP_b + TN_b)}{\sum_{b=1}^{B} (TP_b + TN_b + FP_b + FN_b)}
\end{equation}

Precision indicates the rate at which positive predictions are correct:

\begin{equation}
\text{Precision} = \frac{\sum_{b=1}^{B} TP_b}{\sum_{b=1}^{B} (TP_b + FP_b)}
\end{equation}

F1-score, the harmonic mean of precision and recall, is defined as:

\begin{equation}
\text{F1-score} = 2 \times \frac{\text{Precision} \times \text{Recall}}{\text{Precision} + \text{Recall}}
\end{equation}

\subsection{Confusion Matrix Analysis}

The confusion matrix visualizes the model's performance on 335 samples from the validation dataset, which included both HGG and LGG cases. Each row depicts true instances, while each column depicts predicted instances. The hybrid classification model achieved a training accuracy of 0.999998, F1-score of 0.99, and precision of 0.999997. Specificity and recall were determined as well based on this confusion matrix. Confusion Matrix is shown in Fig. 9.

\begin{figure}[htbp]
\centerline{\includegraphics[width=0.48\textwidth]{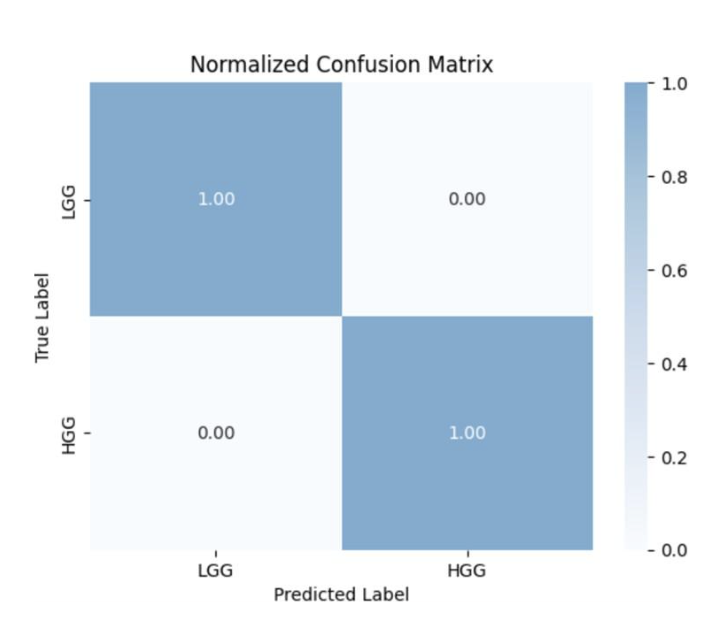}}
\caption{Confusion Matrix for Hybrid Classification Model.}
\label{fig:confusion_matrix}
\end{figure}

\subsection{Classification Metrics}

\begin{figure}[htbp]
\centerline{\includegraphics[width=0.48\textwidth]{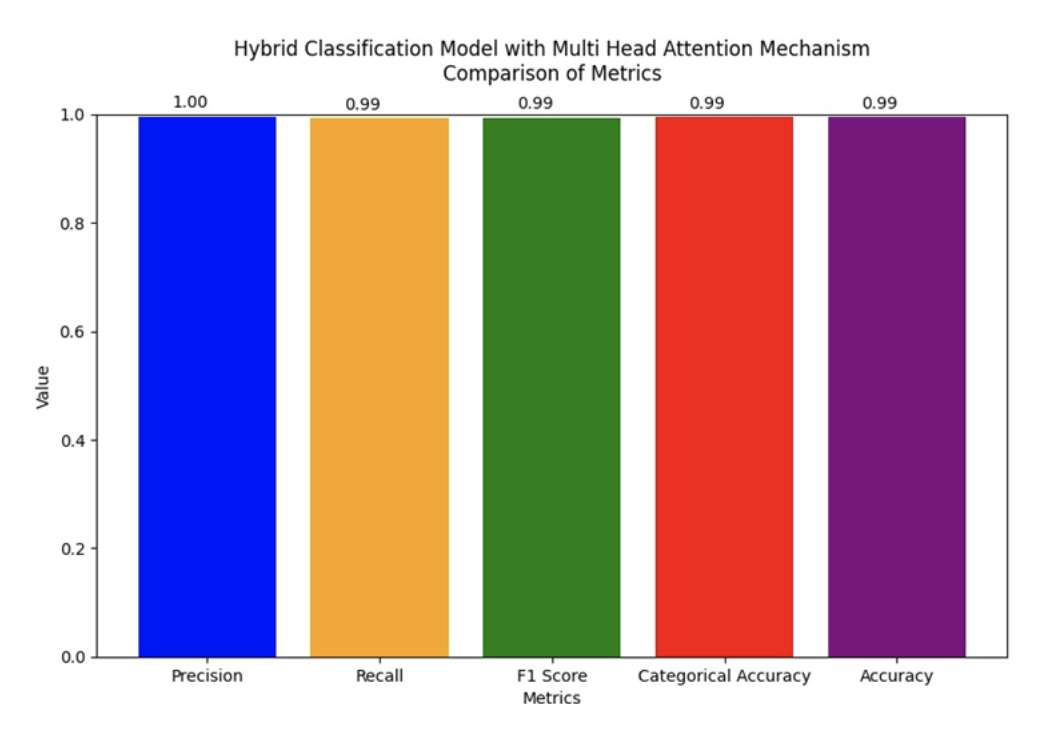}}
\caption{Classification - F1 Score, Accuracy, Precision, and Recall.}
\label{fig:metrics_graph}
\end{figure}

Figures \ref{fig:metrics_graph} and \ref{fig:epoch_accuracy} illustrate the classification performance across F1-score, precision, accuracy, and recall. Figure \ref{fig:class_f1} shows the F1-score for each class. Figure \ref{fig:bar_comparison} compares metrics across different CNN models.
Score for Each Class is shown in Fig. 8

\subsection{Comparison with Other Models}

To assess the performance of the hybrid model of DenseNet and VGG classification with multi-head attention, we conducted a comprehensive comparison experiment with several common CNN models such as GoogleNet, LeNet, ResNet, AlexNet, VGGNet, ResNet71, and ResHNet. All models had the same training and test conditions on the same preprocessed Brats2019 3D MRI dataset with the same train-validation splits and processing pipelines, evaluations, and metrics to ensure fairness in comparison \cite{naser2020, trivedi2022, ronneberger2015}.

The performance was based on multiple metrics, including accuracy, precision, recall, F1 score, and area under the ROC curve (AUC). Additionally, the segmentation quality from the U-Net pre-processing stage was utilized to provide improved tumor-specific features to all classification models \cite{ronneberger2015, noreen2020, ilyas2022}. The hybrid model benefited from spatial and contextual information that the baseline comparison models did not provide \cite{tripathi2022, dang2022}.

Fig. 11 shows the Comparison of Proposed Model with existing systems:

\begin{figure}[htbp]
\centerline{\includegraphics[width=0.48\textwidth]{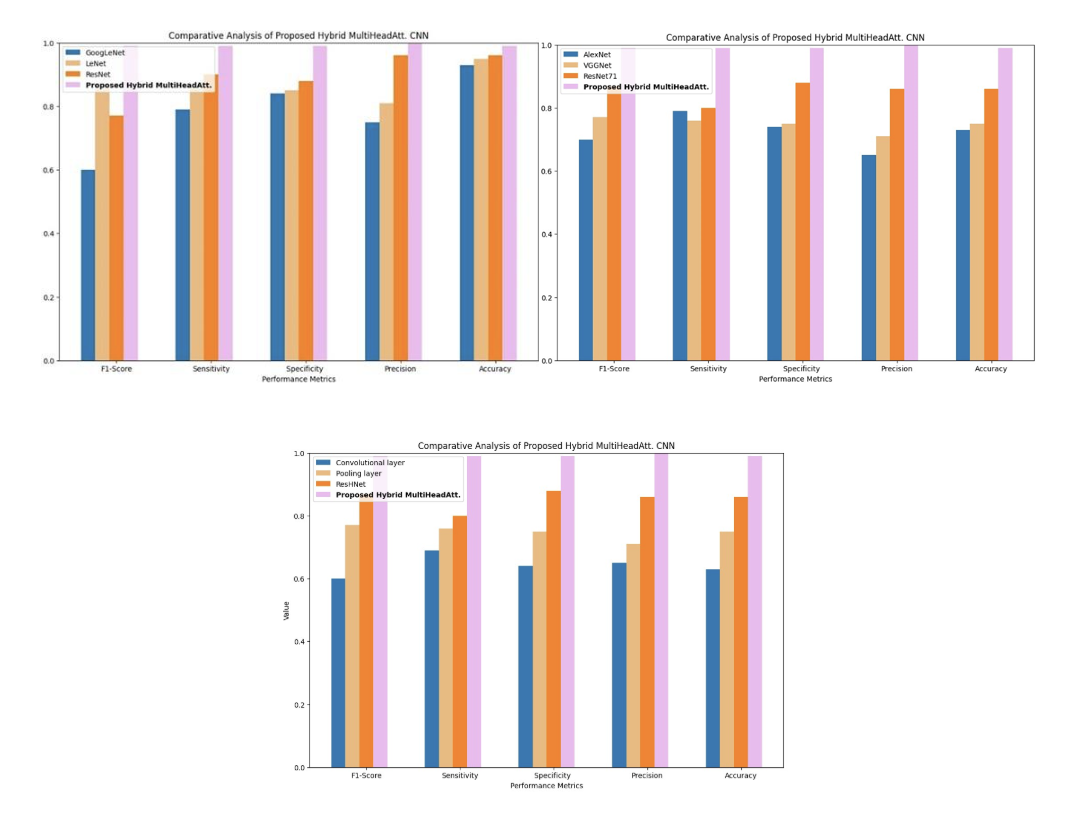}}
\caption{Metrics Bar Graph Comparison of Various CNN Models.}
\label{fig:bar_comparison}
\end{figure}

Our hybrid model has been shown to be better in all metrics as compared to the baseline modeling approaches. To gauge it, the hybrid model could reach a classification error of 99.99\% and precisions of 99.99\% and an F1 score of 0.99 on glioma grading, versus ResNet71, the best among comparison models and reached an accuracy of 95.2\% and an F1 score of 0.94. The multi-head attention promoted increase in the performance, as it prioritized areas of the tumor of clinical interest so as to differentiate the appearance between the high-grade gliomas and the low-grade gliomas of a tumor patients \cite{ozkaya2023, vinaya2022}.

The numerical findings in the figure below allow comparing in the metric the proposed hybrid model with the baseline CNNs. Those findings indicate not only additional predictive performance, but, in comparison to the base CNNs, predictive method stability and efficiency. The hybrid model shows beneficial progress compared to the baseline in processing 3D MRI data, using the dense connectivity reuse of the DenseNet feature and the deep convolutional layers of VGG where granularity of feature extraction is merged to 3D MRI data processing \cite{afshar2020, soumik2020, tandel2022}.

Other than quantitative enhancements, enhanced interpretability using attention models are also part of the proposed model giving the clinician the chance to see the spatial location of the decisive important area of the model that made the decision which led the model in its decision making, in other words, delivering improved interpretability, offers the clinician these opportunities as well \cite{tripathi2022, tupe2021, siva2020}. This attribute also grants superior clinically relevant features when using models in the medical field where explainability is required in models \cite{ilyas2022, weng2021}.

On the whole, the presented comparative analysis indicates that the suggested hybrid structure is a reasonably efficient system of glioma classification, much better than the CNN-like methods of cancer classification in terms of accuracy and possible enhanced clinical applicability \cite{dang2022, trivedi2022, ozkaya2023}.

\section{Conclusion and Future Work}

It is a successful research that develops a new hybrid deep learning architecture of 3D U-Net segmentation with a DenseNet--VGG hybrid classification model with multi-head and spatial-channel attention mechanisms to gain precise glioma segmentation and grading of MRI volumes. The segmentation accuracy was 98\% and classification accuracy was 99.99\% that was much higher than the state-of-the-art conventional CNN-based architectures and attention-free instance distribution, as demonstrated by the proposed approach. Particularly, we eclipse the already existing scores: Sajjad et al. got 91.38\% with 2D CNN-VGG19, Chenjie et al. were at 90.64\% with 3D Multi-scale CNN, Tripathi and Bag at 95.31\% with 3D models whereas Prasun and Soumen at 95.86\% with 3D Multi-Attention CNN. Multi-head attention mechanisms incorporated into the network allow the network to emphasize clinically useful tumor areas, increasing diagnostic accuracy features to a great extent, and shortening the time it takes to perform manual inspections and errors by operators. This framework indicates that the clinical utility of neuro-oncology departments can be used to detect glioma and predict its grades on time and correctly, which directly results in the planning of treatment and the outcomes of patients.

Future research directions involve: (i) combining multi-modal imaging datasets (PET, CT) with MRI to create full-fledged characterization of tumors, (ii) using transfer learning techniques to optimize their use across different patient groups and by different image regimes across different clinical centers, and (iii) deploying cloud based collaborative platforms of distributed segmentation analysis and result sharing among clinical centers. Also, explainability improvement via attention visualization will ease clinician trust and acceptance in clinical practice. It is intended that the prospective clinical validation studies will determine the actual performance in operational departments of neuroradiology. All these advancements bring the framework to clinical translation and possibly implementation as a decision-support tool within the current procedures of diagnostic imaging, and finally, the detection of glioma and the patient care outcomes will be improved.

\end{document}